\documentclass{article}
\usepackage{PRIMEarxiv}
\usepackage{tabularx}

\usepackage[utf8]{inputenc} 
\usepackage[T1]{fontenc}    
\usepackage{hyperref}       
\usepackage{url}            
\usepackage{booktabs}       
\usepackage{amsfonts}       
\usepackage{nicefrac}       
\usepackage{microtype}      
\usepackage{lipsum}
\usepackage{fancyhdr}       
\usepackage{graphicx}       
\graphicspath{{media/}}     
\usepackage{xcolor}         
\usepackage{natbib}
\usepackage{subcaption}
\usepackage{wrapfig}
\usepackage{fvextra}
\usepackage{fancyvrb}
\DefineVerbatimEnvironment{codeblock}{Verbatim}
  {fontsize=\small, breaklines=true, frame=single, numbers=left, numbersep=4pt}
\title{Performative Thinking? The Brittle Correlation Between CoT Length and Problem Complexity}

\author{
\normalfont
\begin{tabularx}{\textwidth}{X X}
\begin{minipage}[t]{0.48\textwidth}
\centering
\textbf{Vardhan Palod} \\
School of Computing and AI \\
Arizona State University, USA \\
\texttt{vpalod@asu.edu} \\[0.8em]
\textbf{Kaya Stechly} \\
Dept. of Computer Science and Wu Tsai Institute \\
Yale University, USA \\
\texttt{kaya.stechly@yale.edu}
\end{minipage}
&
\begin{minipage}[t]{0.48\textwidth}
\centering
\textbf{Karthik Valmeekam} \\
School of Computing and AI \\
Arizona State University, USA \\
\texttt{kvalmeek@asu.edu} \\[0.8em]
\textbf{Subbarao Kambhampati} \\
School of Computing and AI\\
Arizona State University, USA \\
\texttt{rao@asu.edu}
\end{minipage}
\end{tabularx}
}

\begin{document}
\maketitle
\begin{abstract}
Intermediate token generation (ITG), where a model produces output before the solution, has been proposed as a method to improve the performance of language models on reasoning tasks. While these reasoning traces or Chain of Thoughts (CoTs) are correlated with performance gains, the mechanisms underlying them remain unclear. A prevailing assumption in the community has been to anthropomorphize these tokens as “thinking”, treating longer traces as evidence of higher problem-adaptive computation. In this work, we critically examine whether intermediate token sequence length reflects or correlates with problem difficulty. To do so, we train transformer models from scratch on derivational traces of the A* search algorithm, where the number of operations required to solve a maze problem provides a precise and verifiable measure of problem complexity. We first evaluate the models on trivial free-space problems, finding that even for the simplest tasks, they often produce excessively long reasoning traces and sometimes fail to generate a solution. We then systematically evaluate the model on out-of-distribution problems and find that the intermediate token length and ground truth A* trace length only loosely correlate. We notice that the few cases where correlation appears are those where the problems are closer to the training distribution, suggesting that the effect arises from approximate recall rather than genuine problem-adaptive computation. This suggests that the inherent computational complexity of the problem instance is not a significant factor, but rather its distributional distance from the training data. These results challenge the assumption that intermediate trace generation is adaptive to problem difficulty and caution against interpreting longer sequences in systems like R1 as automatically indicative of “thinking effort”.
\end{abstract}
\section{Introduction}
Recent advances in general planning and problem solving have been spearheaded by “Long Chain-of-Thought” models, most notably DeepSeek’s R1 \citep{guo2025deepseek}. These transformer-based large language models undergo the standard stages of pre-training, instruction tuning, and preference alignment, followed by additional post-training on reasoning tasks. At each step, the model is given a question, generates a sequence of intermediate tokens (often called a Chain of Thought or reasoning trace), and ends with a specially delimited answer. After this answer is verified by a formal system, the model’s parameters are updated to increase the likelihood of producing correct final outputs.

Although no optimization pressure is typically applied to the intermediate tokens themselves~\cite{baker2025monitoring, zhou2025r1}, models empirically perform better across many domains when they generate them first~\cite{nye2021show, wei2022chain, zhang2022automatic, hsieh2023distilling, gu2023minillm, guo2025deepseek, pfau2024let, muennighoff2025s1, li2025llms}. While the performance gains are well established, their underlying causes remain unclear. Prior work often interprets these traces anthropomorphically, claiming that models are “thinking” before answering~\cite{nye2021show, gandhi2025cognitive, guo2025deepseek, yang2025understanding, zhou2025r1, bubeck2023sparks}. At the same time, producing longer reasoning traces has been described as \textit{inference-time scaling}—the idea that models perform problem-adaptive computation. Yet there is little reason to expect such adaptivity given how these models are trained. In practice, post-training methods such as supervised fine-tuning with derivational traces or reinforcement learning with rewards based only on final answers are employed \citep{zelikman2022star, li2025llms, muennighoff2025s1, lightman2023let, lambert2024t, yuan2024free, guo2025deepseek, sun2023reinforcement, arora2502training, yu2025dapo}. In both cases, models are prompted to generate intermediate tokens before producing their answers, but optimization pressure applies only to the correctness of the final output. The intermediate tokens, being samples from the model’s base policy, are not explicitly aligned with problem difficulty, correctness, or structured reasoning. Consequently, these traces are not guaranteed to reflect problem-adaptive computation and may instead be arbitrary byproducts of the generative process.

Following previous work that elucidated important functional aspects of large-scale models through controlled small-scale experiments \cite{wang2024grokked, power2022grokking, zhong2023clock} and working within a sort of “model organism” paradigm, we focus on fully controlled, open, and replicable models trained from scratch. Specifically, we train transformer models from scratch on derivational traces of the A* search algorithm. This approach offers two advantages: (1) the reasoning procedure is explicitly defined and verifiable, and (2) the complexity of input problems can be precisely manipulated.

Within this framework, we evaluate whether intermediate token length in transformer-based models really reflects problem difficulty. We first evaluate the model on the simplest path-finding tasks: free-space problems without obstacles. Solving these should, in principle, require minimal computation, yet we find that the model often performs excessively long computations, and in many cases even fails to produce a solution altogether, thus calling into question the interpretation that the length of intermediate tokens correlates with problem complexity. We then evaluate across diverse maze-generation algorithms. On distributions close to the training distribution, intermediate token lengths show some alignment with problem difficulty. However, this correlation vanishes on more structurally distinct distributions, where trace length and problem complexity become entirely decoupled. These results suggest that the apparent adaptivity of intermediate token length is not evidence of problem-sensitive computation. Instead, it may largely reflect the approximate recall from the training distribution, rather than any genuine alignment between computation length and problem difficulty. Our findings thus suggest that the commonly assumed link between “thinking time” (as measured by reasoning trace length) and task difficulty is misleading.

\section{Related Works}
\textbf{Post-Training for Reasoning -}
Recent progress in improving the reasoning capabilities of Large Language Models (LLMs) has been driven by methods that train models not only on correct answers, but also on reasoning traces that lead to those answers \citep{zelikman2022star, li2025llms, muennighoff2025s1, lightman2023let, lambert2024t, yuan2024free, sun2023reinforcement, guo2025deepseek, arora2502training, yu2025dapo}. Whether this is achieved via supervised fine-tuning on trace-augmented datasets or via reinforcement learning techniques like Group Relative Policy Optimization (GRPO), the end result is the same: models "think" for varying amounts of time by outputting additional tokens before outputting their final answers, and performance measures improve.
These methods are often paired with claims about how and why they work, which lean on anthropomorphic framings of the intermediate tokens as model "thinking". These claims often treat the length of the intermediate traces as the amount of "thinking" a model does before producing a final answer. Thus, there is an implicit assumption that the traces will be problem-adaptive, i.e., easier problems will have shorter traces and harder problems will have longer traces.

\textbf{Training Transformers on Traces -}
Prior efforts have trained models from scratch to mimic search algorithms like A*, Breadth-First Search (BFS), and Monte Carlo Tree Search for tasks in pathfinding, arithmetic, and general problem-solving \citep{lehnert2024beyond, gandhi2024stream, yang2022chain}, as well as the DPLL procedure for Boolean SAT problems \citep{pan2024can}. However, while these studies train on traces of formal procedures, we believe no other work has analyzed whether there is any correlation between the trained model's traces and the ground truth problem complexity.

\section{Methodology}
We study a standard grid-based pathfinding domain: finding a legal path from a start to a goal cell in a $30 \times 30$ grid where each cell is either free or a wall. The agent begins at the start state and can move up, down, left, or right. The transformer receives a tokenized description of the problem \cite{lehnert2024beyond,su2024dualformer} and must output a plan—a sequence of actions that moves the agent only through free cells and terminates at the goal.

Navigation problems are generated using diverse maze-generation algorithms (Appendix \ref{Maze:gen_algo}), producing varied structural patterns. Following Searchformer and Stream of Search \cite{lehnert2024beyond,gandhi2024stream}, we modify A* to emit a linearized trace: each node creation is logged as create x y cG cH and each expansion as close x y cG cH, where $cG$ is the cost from start ($g(n)$) and $cH$ is the heuristic estimate to goal ($h(n)$). In our experiments, we formalize problem difficulty as the number of operations the A* algorithm takes to solve the problem. Thus, we consider that the problem difficulty increases as the number of operations A* takes to solve the problem. Note that the trace length of A* is directly proportional to the number of operations the algorithm takes to solve the problem. An example of a problem, its A* trace, and solution is given in Appendix \ref{sec:example}.

To construct our training sets, we generate 500,000 mazes using Wilson’s algorithm and randomly select a start and goal cell. Then, we use A* with the Manhattan distance heuristic to find an optimal plan for each maze as well as to produce a trace that is saved with each datapoint. We modify the architecture of the Qwen2.5 0.5B~\cite{qwen2.5_2024} to support a vocabulary of exactly 944 different tokens (which reduces the parameter count to about 380 million from 500 million), randomly initialize the model, and then train it for 255,000 training steps with a batch size of 8 on two NVIDIA H100s. Note that we are training empty transformer models from scratch instead of fine-tuning pre-trained models. To assess whether the model has learned to solve pathfinding problems, we first evaluate it on 1,000 held-out Wilson mazes, where it achieves 80\% accuracy in generating correct plans. We also evaluate this model on held-out problems generated by distinct maze-generation algorithms: Free-space, Kruskal, DFS, SF-Style, and Drunkard. Additional training details are given in Appendix \ref{model:training}.

\section{Results}
\subsection{Excessive Intermediate token length for the easiest problems}
To probe the model’s behavior on the simplest possible tasks, we evaluated it on 100 \textit{free-space} problems, where the grid contains no obstacles and the agent must simply navigate from the start state to the goal. More details on the generation of these problems are provided in the Appendix.

\begin{figure}[ht]
    \centering
    \begin{subfigure}[t]{0.40\linewidth}
        \centering
        \includegraphics[height=4cm]{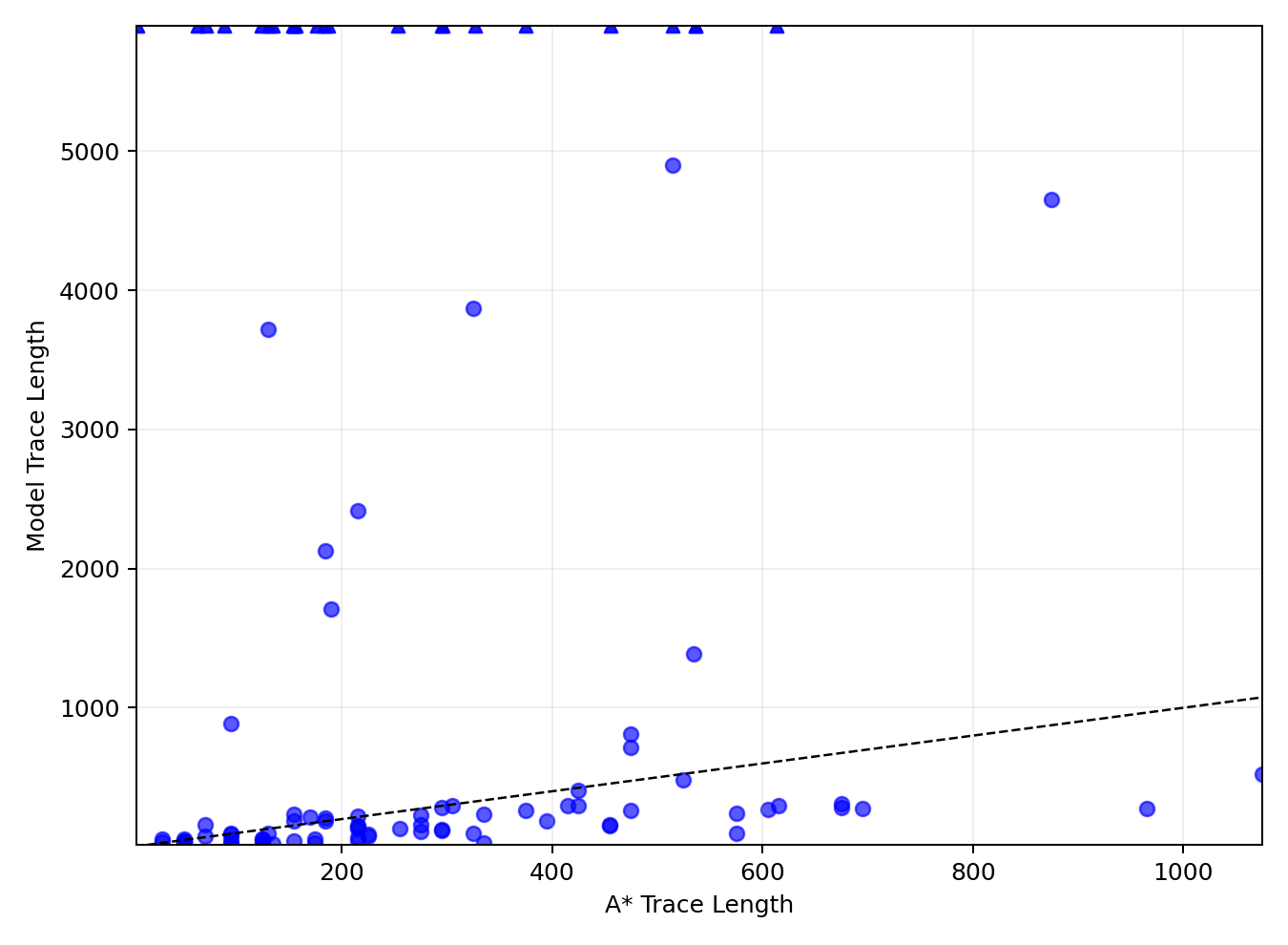}
        \caption{Responses on \textit{Free-space problems}. Points at the top correspond to failures where the model generated tokens until reaching the 32k limit.}
        \label{fig:correct_model_responses_free_space}
    \end{subfigure}
    \hspace{1.5cm}
    \begin{subfigure}[t]{0.40\linewidth}
        \centering
        \includegraphics[height=4cm]{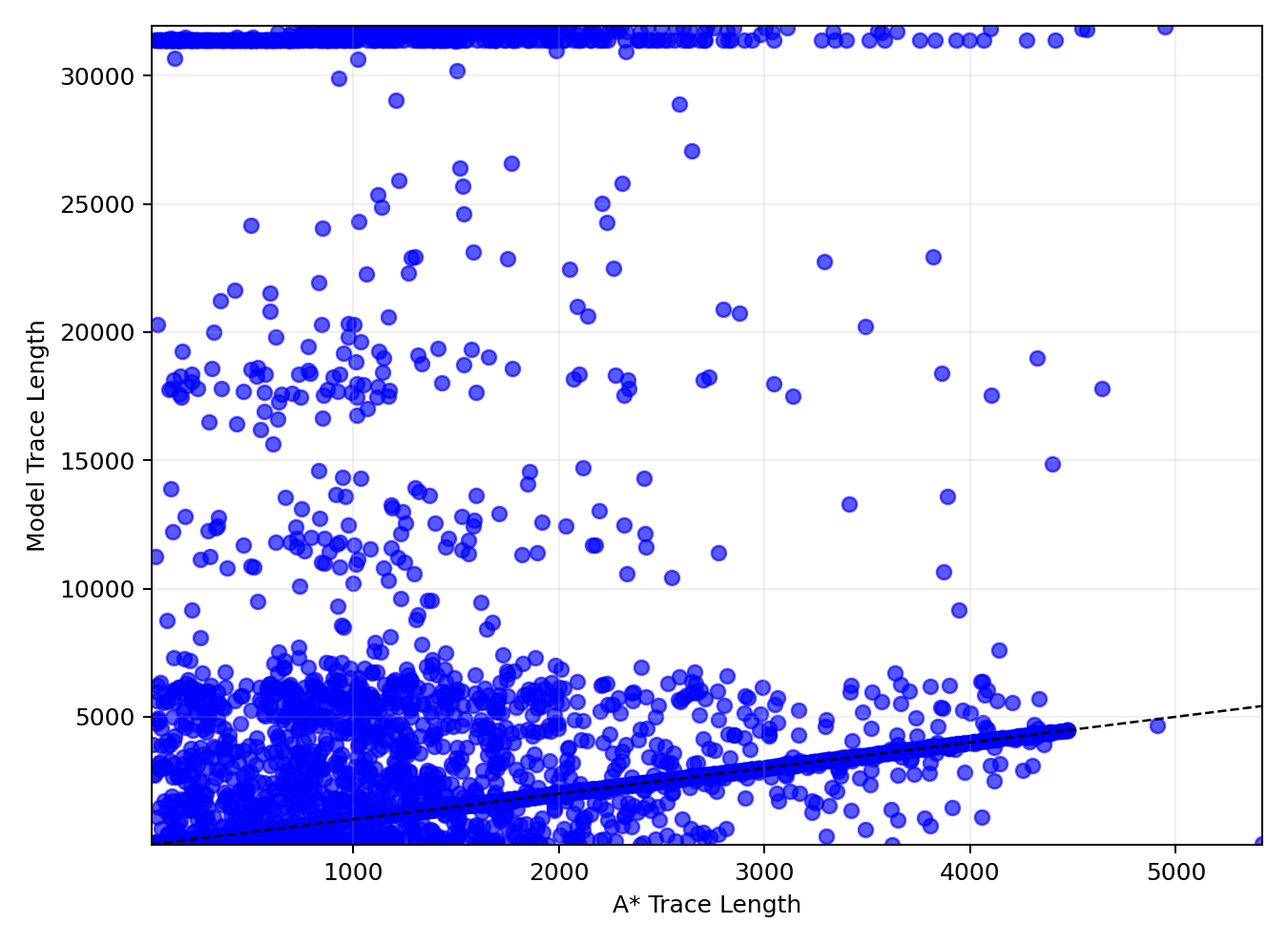}
        \caption{Responses on problems generated with DFS, Drunkard, Searchformer-style, Wilson, and Kruskal algorithms (1000 each).}
        \label{fig:correct_model_responses}
    \end{subfigure}
    \caption{Comparison of generated intermediate token length (Y-axis) with ground-truth A* trace length (X-axis). The dashed line represents $y=x$.}
    \label{fig:trace_length_comparison}
\end{figure}

Even under this trivial setting, the model performs poorly: only 5 out of 100 problems yield a valid plan. More strikingly, as shown in Figure~\ref{fig:correct_model_responses_free_space}, there is very little correlation between the model’s intermediate token length and the ground-truth A* trace length, as almost no responses lie on or near the y = x line. If there were a correlation, the generated trace lengths would closely match the ground-truth values. In fact, on some of the easiest problems, the model continues producing tokens up to the maximum context limit of 32k without ever producing a valid solution.

\subsection{Analysis of responses on problems generated by various algorithms}
We further evaluate the model, trained on Wilson-generated problems, on held-out instances produced by five distinct maze-generation algorithms: Wilson, Kruskal, DFS, SF-Style, and Drunkard. As shown in Figure~\ref{fig:correct_model_responses}, the core finding persists: the amount of computation, as reflected in the length of intermediate traces, remains largely agnostic to problem complexity. The scatter plots are broadly populated across the range of problem difficulties, indicating only a very loose correlation between the model’s intermediate trace lengths and the ground-truth A* trace lengths.

\textbf{Analyzing In-Distribution vs.\ Out-of-Distribution Responses}

\begin{figure}[ht!]
    \centering
    \begin{subfigure}[t]{0.40\linewidth}
        \centering
        \includegraphics[height=4cm]{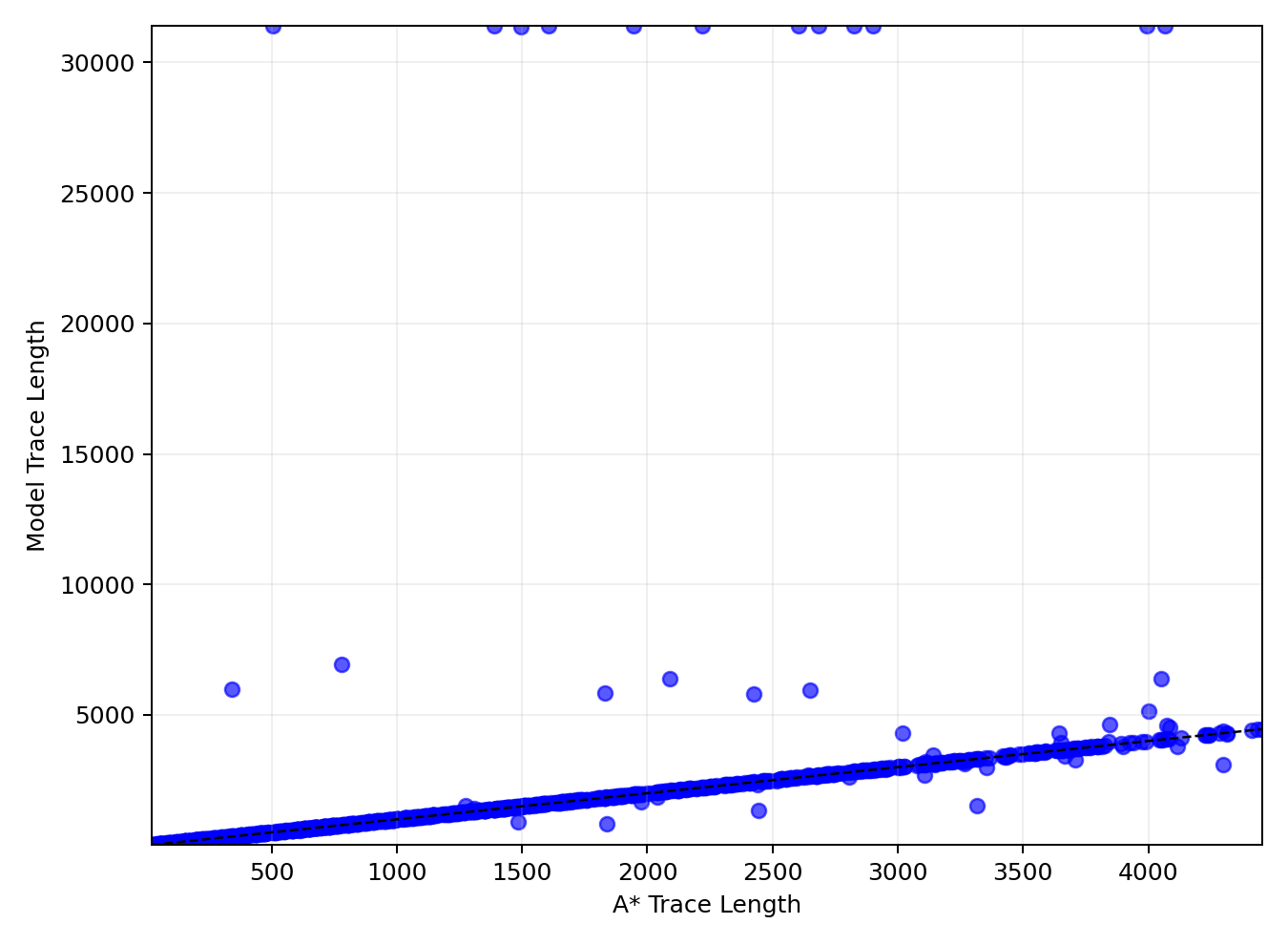}
        \caption{Wilson trace length scatter}
        \label{fig:wilson_trace}
    \end{subfigure}
    \hspace{1.5cm}
    \begin{subfigure}[t]{0.4\linewidth}
        \centering
        \includegraphics[height=4cm]{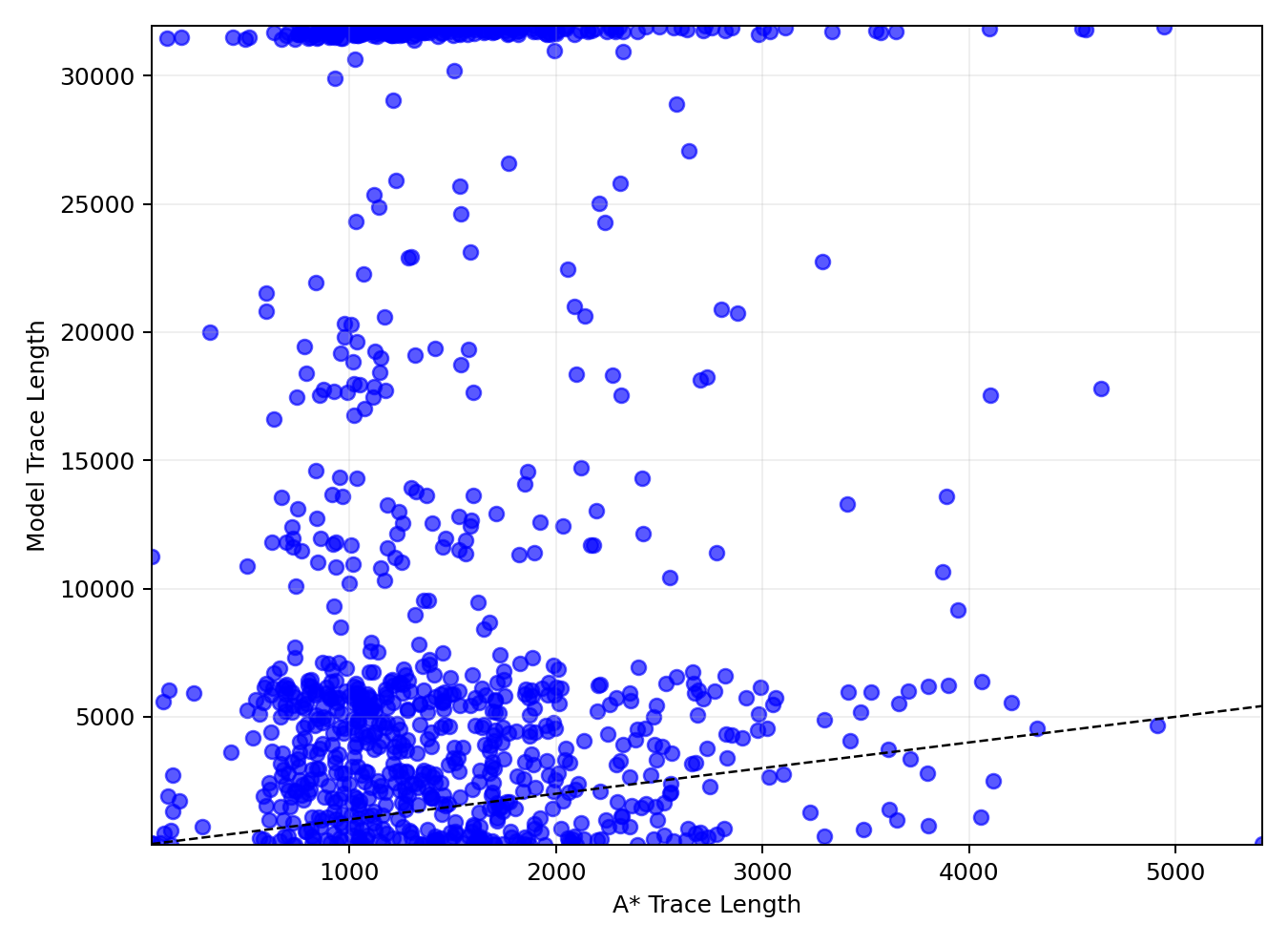}
        \caption{Searchformer trace length scatter}
        \label{fig:searchformer_trace}
    \end{subfigure}
    \caption{Comparison of trace length scatter plots for problems generated using Wilson and Searchformer Algorithms.}
    \label{fig:trace_length_scatter}
\end{figure}

We also evaluate differences in the model’s behavior on held-out in-distribution problems compared to out-of-distribution problems.
Since the model is trained on Wilson-generated mazes, it learns to approximate A* traces in those cases, yielding a visible alignment between intermediate token lengths and ground-truth trace lengths (Fig. \ref{fig:wilson_trace}). However, when evaluated on Searchformer-style mazes \cite{lehnert2024beyond}, this correlation disappears entirely (Fig. \ref{fig:searchformer_trace}). In such cases, the intermediate trace lengths fluctuate independently of problem complexity. Together, these results suggest that the apparent problem-adaptive computation observed under Wilson mazes is due to the test set problem instances, even though held out, being drawn from the same distribution as the training data. As the distributional distance increases—as in the case of Searchformer-style mazes—the correlation between trace length and problem complexity vanishes.

\section{Discussion}
While intermediate token generation has often been interpreted as reflecting problem-adaptive computation, our results suggest that this assumption is misleading. When the distributional distance between test and training data is small, as in the case of Wilson mazes, intermediate token length may appear correlated with problem complexity. However, as this distance increases, such as with Searchformer-style or Drunkard mazes, the correlation vanishes. These observations carry implications for how reasoning tokens are interpreted. Broadly, our results indicate that trace length is not a function of the from-scratch computational complexity of the problem instance, but rather of the distributional distance between that instance and the instances that the LLM/LRM has been trained on. Since intermediate token generation does not reliably indicate “thinking” or the computational effort expended to solve a problem, methods that claim to improve efficiency in large reasoning models by reducing reasoning-chain length may be based on a flawed premise~\cite{Yuan2025EfficientRTA, arora2502training, Chen2025VeriThinkerLTA, Hassid2025DontOIA, Liu2025ThoughtMEA, Li2025AALCLLA, shrivastava2025samplethinklessgroup}.

\section{Conclusion}
In this work, we analyzed whether the length of intermediate tokens reflects problem-adaptive computation in transformer-based large reasoning models (LRMs). Using a controlled setting with transformers trained from scratch on verifiable A* traces, we demonstrated that intermediate token length and problem complexity are loosely correlated.

Our findings suggest that the commonly assumed link between “thinking time” (as measured by reasoning trace length) and task difficulty is misleading. More broadly, our results caution against anthropomorphizing intermediate reasoning tokens and highlight the need for more rigorous methodologies when interpreting the internal processes of transformer-based reasoning models.

\newpage
\bibliography{references, sf_references}
\bibliographystyle{plain}

\newpage
\appendix
\section{Appendix}
\subsection{Maze Generation Algorithms}
\label{Maze:gen_algo}
We generate navigation problems using diverse generation algorithms, resulting in varied structural patterns and exploration dynamics. This enables systematic out-of-distribution (OOD) evaluation by testing models on maze types unseen during training -- which was all done on mazes generated with Wilson's algorithm. These generation algorithms can be sorted into two major categories: 1) algorithms that do not permit cycles and sample over the spanning trees of the $30\times30$ grid, and 2) algorithms that permit loops and create noisy, less-structured dungeon or cave-like instances. For all algorithms except Searchformer's, which has its own start and goal generation loop, we sample a legal (start, goal) pair after maze generation.  We also generate problems with no obstacles in the grid called free-space problems.

\textbf{Acyclic Maze Generation}
\begin{enumerate}
    \item \textbf{Wilson's algorithm:} This is the algorithm that we use to generate mazes for training models. Wilson's algorithm generates uniform random mazes by performing loop-erased random walks from unvisited cells until they connect to the current maze~\citep{wilson1996generating}. Each walk removes any loops it creates, ensuring a valid tree structure. This process continues until all cells are included, producing a uniform sample from the space of all possible spanning trees of the $30\times30$ graph.

    \item \textbf{Kruskal's algorithm:} Kruskal’s algorithm, originally proposed for finding a minimum spanning forest of an undirected edge-weighted graph \citep{kruskal1956shortest}, generates mazes by treating each cell as a node and randomly removing walls between unconnected regions, using a union–find structure to avoid cycles. This results in a fully connected maze without loops, though the maze distribution is not perfectly uniform. The method produces mazes biased towards short local connections and dead ends.
    
    \item \textbf{Randomized Depth-First Search algorithm:} The randomized depth-first search (DFS) or recursive backtracker algorithm generates mazes by carving a path forward until reaching a dead-end \citep{tarjan1972depth}. When it hits a dead-end (no unvisited neighbors), it backtracks until it finds a new direction to explore, repeating until all cells are visited and connected into a complete maze. Depth-first search is biased towards generating mazes with low branching factors and many long corridors.
\end{enumerate}

\textbf{Cave Generation}

\begin{enumerate}
    \item[4.] \textbf{Drunkard's Walk:} We implement a version of the “Drunkard's Walk” algorithm, as described by \cite{jrheard2016drunkards}, and originally used for procedurally generating dungeons for top-down two-dimensional video games. Starting from a grid of solid walls, a random walk is performed, carving out the current cell on every step. The walk continues until a predefined number or percentage of floor tiles has been dug out. This method preserves cycles, producing cave-like structures with open chambers and looping corridors. The output space includes grid states unreachable by perfect acyclic maze generators.

    \item[5.] \textbf{Searchformer style generation:} We also implement the random generation algorithm used in the Searchformer paper \cite{lehnert2024beyond}, though we use it for evaluation rather than training. Tasks are generated by exhaustive rejection sampling: first, randomly select a number between 30\% and 50\%. Then select that percentage of cells to be wall cells. Randomly choose a start and goal location and execute A$^*$ to find an optimal plan. Reject unsolvable, too easy, or duplicate instances and resample. These instances also allow for loops and so are also out of distribution for our models.
\end{enumerate}

\textbf{No-Obstacles Generation}  
\begin{enumerate}
    \item[6.] \textbf{Free-space algorithm:} In this algorithm, we first select the number of levels of outer walls. In our case, we chose the number of levels of outer walls to be 4. Within the inner grid, a start cell and a goal cell are then chosen randomly, with no walls placed in the inner grid. Finally, we execute A$^*$ to generate the intermediate trace and obtain the optimal plan.
\end{enumerate}

\subsection{Model Training and experimental details}
\label{model:training}
For all experiments, we trained decoder-only Qwen-2.5-0.5B models, customized with a domain-specific tokenizer that reduced the effective parameter count to approximately 380M. The models were randomly initialized and trained for 255,000 steps with an effective batch size of 8 on two NVIDIA H100 GPUs. The model has a context length of 32,000 tokens to support the long lengths of intermediate token generation. Training was performed on a dataset of 500k Wilson generated problems, after which the models were evaluated on 1,000 held-out instances generated by five distinct maze generation algorithms: Wilson, Kruskal, DFS, SF-Style, and Drunkard. 

We optimized with AdamW ($\beta_1$=0.9, $\beta_2$=0.999) and applied a weight decay of 0.1528, a peak learning rate of 2.2758e-4, and 100 warm-up steps. All experiments were run under bf16 precision with fixed random seeds for reproducibility. 

\subsection{Example Problem Instance}\label{sec:example}
\begin{figure}[tbp]
    \centering
    \includegraphics[height=5cm]{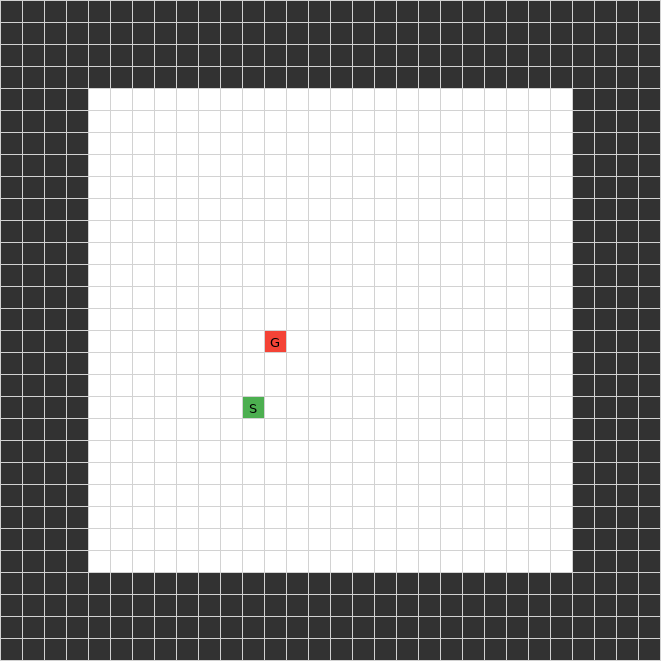}
    \caption{Example grid of a Free-space problem. It is a 30x30 grid with no obstacles in the inner grid. The start and goal states are chosen randomly.}
    \label{fig:eg_grid}
\end{figure}

Problem description, A* trace, and solution are given in tokenized form, following \cite{lehnert2024beyond, gandhi2024stream}. The visualization of the problem is given in figure \ref{fig:eg_grid}. 

\paragraph{Problem}
\begin{codeblock}
start 18 11 goal 15 12 wall 0 0 wall 0 1 wall 0 2 wall 0 3 wall 0 4 wall 0 5
wall 0 6 wall 0 7 wall 0 8 wall 0 9 wall 0 10 wall 0 11 wall 0 12 wall 0 13 ..
\end{codeblock}

\paragraph{A* search trace}
\begin{codeblock}
close 18 11 c0 c4 create 17 11 c1 c3 create 19 11 c1 c5 create 18 10 c1 c5
create 18 12 c1 c3 close 17 11 c1 c3 create 16 11 c2 c2 create 17 10 c2 c4
create 17 12 c2 c2 close 18 12 c1 c3 create 19 12 c2 c4 create 18 13 c2 c4
close 16 11 c2 c2 create 15 11 c3 c1 create 16 10 c3 c3 create 16 12 c3 c1
close 17 12 c2 c2 create 17 13 c3 c3 close 15 11 c3 c1 create 14 11 c4 c2
create 15 10 c4 c2 create 15 12 c4 c0 close 16 12 c3 c1 create 16 13 c4 c2
close 15 12 c4 c0
\end{codeblock}

\paragraph{Solution}
\begin{codeblock}
plan 18 11 plan 17 11 plan 16 11 plan 15 11 plan 15 12
\end{codeblock}

\end{document}